\documentclass[letterpaper, 10 pt, conference]{ieeeconf}

\IEEEoverridecommandlockouts                              

\usepackage{amsmath, amsfonts, bm}
\usepackage{graphicx}
\usepackage{tabularx}
\usepackage{graphics}
\usepackage[usenames,dvipsnames]{color}
\usepackage[dvipsnames]{xcolor}
\usepackage{booktabs}
\usepackage{hyperref}

\usepackage{subcaption}

\newcolumntype{C}[1]{>{\centering\arraybackslash}p{#1}}
\newcommand{\vcentered}[1]{\begin{tabular}{c} \hspace{-15pt} #1 \end{tabular}}

\newcommand{\Tshearside}{\bm{T}_{\text{sideway}}}
\newcommand{\Tshearforward}{\bm{T}_{\text{forward}}}

\newcommand{\Tscale}{\bm{T}_{\text{scale}}}
\newcommand{\Tflip}{\bm{T}_{\text{flip}}}

\newcommand{\distanceGeodesic}{\bm{D}^g}
\newcommand{\distanceGeodesicUpdated}{\bm{D}^{g^*}}
\newcommand{\distanceEuclidean}{\bm{D}^e}
\newcommand{\distanceEuclideanUpdated}{\bm{D}^{e^*}}
\newcommand{\distanceGaussian}{\bm{D}}
\newcommand{\distanceGaussianUpdated}{\bm{D}^{*}}

\newcommand{\shape}{\bm{P}}
\newcommand{\shapeUpdated}{\bm{P}^*}
\newcommand{\point}{\bm{p}}
\newcommand{\keypoints}{\bm{N}}
\newcommand{\keypointsUpdated}{\bm{N}^*}
\newcommand{\keypoint}{\bm{n}}

\newcommand{\numberOfCameras}{C}
\newcommand{\cameraID}{c}
\newcommand{\cameraDropProbability}{0.95}

\newcommand{\numberOfPoints}{n}
\newcommand{\numberOfKeypoints}{m}

\title{\LARGE \bf
Semantic keypoint extraction for scanned animals\\using multi-depth-camera systems
}

\author{Raphael Falque$^{1}$, Teresa Vidal-Calleja$^{1}$, and Alen Alempijevic$^{1}$
\thanks{$^{1}$The authors are with the Robotics Institute,
        University of Technology Sidney, Australia
        {\tt\small raphael.guenot-falque@uts.edu.au}}
}

\begin{document}
\maketitle

\begin{abstract}
    Keypoint annotation in point clouds is an important task for 3D reconstruction, object tracking and alignment, in particular in deformable or moving scenes. In the context of agriculture robotics, it is a critical task for livestock automation to work toward condition assessment or behaviour recognition. In this work, we propose a novel approach for semantic keypoint annotation in point clouds, by reformulating the keypoint extraction as a regression problem of the distance between the keypoints and the rest of the point cloud. We use the distance on the point cloud manifold mapped into a radial basis function (RBF), which is then learned using an encoder-decoder architecture. Special consideration is given to the data augmentation specific to multi-depth-camera systems by considering noise over the extrinsic calibration and camera frame dropout. Additionally, we investigate computationally efficient non-rigid deformation methods that can be applied to animal point clouds. Our method is tested on data collected in the field, on moving beef cattle, with a calibrated system of multiple hardware-synchronised RGB-D cameras.
    
\end{abstract}



\section{Introduction}


Point clouds are a common data representation generated by robotics perception systems equipped with depth sensors, such as stereo/depth cameras and LiDARs. Complex perception systems integrate several depth sensors allowing them to maximise the coverage of their surroundings~\cite{heng2019project, brandao2020placing}. For frameworks using multiple depth sensors, it is common to align and merge the point clouds using extrinsic calibration within the front-end data processing. The analysis of the merged point cloud (e.g., semantic segmentation~\cite{behley2019semantickitti}, or scene classification~\cite{ren2022benchmarking}) is then performed as part of what we call back-end process, e.g., through deep learning models for real-time processing. Most deep learning models for point cloud analysis developed within the computer vision and computer graphics communities often overlook the noise generated during the data collection. However, this topic is at the core of robotics applications where both front-end and back-end processes have an impact on the model predictions accuracy and need to be considered. This paper studies how robustness to real-world data collection, such as calibration noise and camera frame dropping, can be built within the training of deep learning models for keypoint extraction in point clouds.

Keypoints extraction and matching is an important task of 3D scene understanding for robotics applications such as tracking, shape retrieval, and registration. For instance, a direct application can be found for non-rigid deformation algorithms which require a good prior in order to solve the alignment of a template and a target as a gradient descent optimization. This surprisingly specific problem formulation finds applications in medical robotics for organs alignment~\cite{zhang2020deep}, clothing manipulation~\cite{huang2022mesh}, humans~\cite{loper2015smpl} and animals~\cite{ruegg2022barc} body shape reconstruction. Furthermore, given a set of keypoints matching pairs, the rigid alignment of two shapes can be solved in a closed-form with singular value decomposition (SVD)~\cite{sorkine2017least} as an alternative to iterative closest point (ICP) methods~\cite{rusinkiewicz2001efficient}.

In the livestock industry, the automation of animal health assessment and monitoring is critical to ensure animal well-being and improved farming productivity. Working toward this direction, automatic pipelines are required to capture and identify key areas in animal bodies~\cite{gardiner2020,kang2021review,qiao2021,mcphee2017live,bercovich2013development}. Real-time 3D shape and model reconstruction have been identified as crucial steps towards explicit segmentation of animal body regions, which are meaningful for cattle posture or behaviour recognition~\cite{qiao2021}. We consider the challenge of extracting keypoints of cattle bodies using a perception system containing synchronised RGB-D cameras capturing multiple point clouds for each animal, shown in Figure~\ref{fig:data_collection}. 

\begin{figure}
    \centering
    \includegraphics[width=\linewidth]{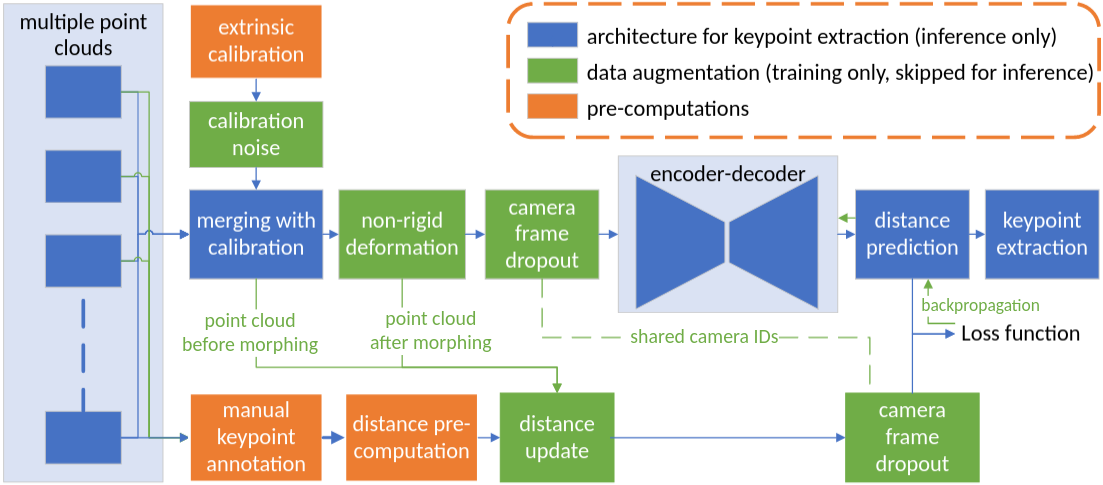}
    \caption{Method overview: at inference time, the point clouds are merged and passed into an encoder-decoder architecture such as Pointnet++~\cite{qi2017pointnet++} or KPConv~\cite{thomas2019kpconv} to extract the keypoints (in blue). During training, a dataset is first manually annotated and the distance on the manifold is pre-computed (in orange). The weights of the encoder-decoder are then trained by using augmentation on the calibration, non-rigid deformation, and camera frame dropout (in green). In our experiments, the encoder-decoder inputs are $\numberOfPoints\times3$ points and the outputs are the $\numberOfPoints\times6$ distances to the $6$ keypoints.}
    \label{fig:encoder-decoder}
\end{figure}

We formulate the semantic keypoints extraction from point clouds as a regression problem with a segmentation encoder-decoder architecture used as the main building block~\cite{qi2017pointnet++,thomas2019kpconv}. In contrast with \cite{ge2018point}, where an encoder-decoder is stacked into an MLP to solve the regression problem of keypoint extraction, we do not modify the encoder-decoder architecture and formulate the regression problem by redefining how the mean squared error (MSE) loss function is computed. A bloc diagram of the method is shown in Figure~\ref{fig:encoder-decoder}. As an additional focus of this paper, we investigate how data augmentation can be adapted to our problem by considering the integration of augmentation specific to multi-depth-camera data and non-rigid deformation formulated in the form of shear transformation matrices. 

The main contributions are: (1) we formulate the keypoint annotation problem of point clouds as a regression of the distance on the manifold mapped into an RBF function. The regression of the distance on the manifold is learned with en encoder-decoder architecture which essentially solves the problem of keypoint matching as a semantic problem (similarly to segmentation problems). We argue that any encoder-decoder model can be used as long as the architecture is able to learn on point clouds data. (2) We consider a complex data augmentation process that includes noise inherent to multi-depth-camera systems, through the efficient use of non-rigid deformation methods for point clouds, and a computationally efficient approximation of the distance on the manifold.

The implementation is available at \href{https://github.com/rFalque/semantic\_keypoints\_extraction}{https://github.com/\\rFalque/semantic\_keypoints\_extraction}.

\section{Related work} 
Pose and body shape estimation from 2D images is a well-studied topic thanks to the large datasets available on the internet (such as COCO \cite{coco2014}). For humans, the pose can be estimated with a deep neural network and the shape can then be optimised over the skeleton pose~\cite{loper2015smpl, osman2020star}. Similar methods have been applied to various animals~\cite{Zuffi:CVPR:2017} and refined for specific breeds such as lions~\cite{Zuffi:CVPR:2018} and zebras~\cite{Zuffi:ICCV:2019}. For 2D images, large datasets with ground truth poses are available, making the use of deep learning applications possible~\cite{ng2022animal}. In cases where the ground trough is not available, 2D images and annotations can still be generated from synthetics animated animals~\cite{mu2020learning}, however, these datasets have a lack of variety with respect to the 3D models of the animals.

In the field of computer graphics, the problem of shape correspondences~\cite{van2011survey} studies the point-wise association between different shapes (generally defined as triangular meshes) and therefore is a generalisation of the keypoints annotation problems. Amongst the most significant work, the study of the shape spectral decomposition and the spectral correspondence has been formulated in a framework called \emph{functional maps}~\cite{ovsjanikov2012functional}. This work has later been extended for partial shapes~\cite{rodola2017partial}, and deep learning approaches have been developed to leverage the spectral domain in a learning context~\cite{litany2017deep, attaiki2021dpfm}.

However, for robotics applications, where LIDAR and RGB-D cameras are used for scanning the environment, working with meshes is often a cumbersome task. Standard mesh reconstruction methods~\cite{bernardini1999ball, kazhdan2013screened} perform poorly on irregular, noisy, and non-closed point clouds. As a result, working directly on point clouds and bypassing the mesh reconstruction is often preferable. While functional maps could be adapted to work directly on point clouds, thanks to robust Laplacian operator~\cite{Sharp:2020:LNT}, they lack robustness (e.g., functional maps can have catastrophic failures) and are over-engineered for the problem of keypoints correspondences (i.e., extracting few keypoints is a significantly simpler task than solving the full shape correspondence).

In the case of keypoint annotation of 3D point clouds, research is sparser due to the complexity of gathering and annotating data. Traditional methods usually consider the keypoints extraction from point cloud data by extracting a 3D skeleton~\cite{tagliasacchi2009curve, cao2010point, tagliasacchi2012mean, lin2021point2skeleton}, from which voting systems can establish correspondences~\cite{au2010electors}. Deep learning methods have recently been developed to predict the 3D skeleton from point clouds~\cite{lin2021point2skeleton}.

Most similar to our method is the recent work on unsupervised keypoint annotations from point clouds using deep learning models. Architectures for learning on point clouds are constantly evolving; the most widespread architectures developed over the last years are Pointnet~\cite{qi2017pointnet}, Pointnet++~\cite{qi2017pointnet++}, and Kp-CONV~\cite{thomas2019kpconv}. \cite{ge2018point} used Pointnet stacked with additional MLP used to transform the Pointnet output into a regression problem. \cite{you2020keypointnet} focus on the loss functions capable of handling multiple human annotations and consider the keypoint extraction as a classification problem. \cite{fernandez2020unsupervised} propose a different architecture that builds upon Pointnet and solves the problem of keypoint annotation as an unsupervised problem using symmetric linear basis shapes. Similarly, \cite{shi2021skeleton} considers the unsupervised problem. They use a Pointnet++ encoder and a custom-made decoder that generates the skeleton from the keypoints, which are then aligned between instances.


\section{Methodology}
\subsection{Problem Definition}
Given the partial 3D shape of an animal represented as a set of point clouds captured by $\numberOfCameras$ depth-cameras $\{\shape_1, \dots, \shape_\numberOfCameras\}$ concatenated into a single point cloud such as $\shape = \bigcup\limits_{\cameraID=1}^{\numberOfCameras} \shape_\cameraID$ where $\shape = \left\{\point_1, ..., \point_\numberOfPoints\right\}$ and $\point_i \in \mathbb{R}^3$, we aim at estimating the $\numberOfKeypoints$ ordered keypoints $\keypoints= \left\{\keypoint_1, ..., \keypoint_\numberOfKeypoints\right\}$, where $\keypoint_j \in \shape$. The keypoints are ordered in the sense that they are systematically annotated in the same order, such as $\{$right rear leg, right front leg, hip, ...$\}$. Therefore, the keypoints have a semantic meaning.

The proposed methodology is summarized in Figure~\ref{fig:encoder-decoder}, and each step is formally defined in this section.

\subsection{Architecture and loss function}
\label{sec:architecture}
We propose to solve the keypoint extraction as a regression problem, where we learn the distance between each point from the point cloud to each annotated point, resulting in $\numberOfKeypoints$ distance vectors of size $\numberOfPoints$. Given this distance, the keypoints could then be obtained by taking the argument of the distance vectors minimum.

From a machine learning perspective, we formulate our problem as a mapping between the $\numberOfPoints \times 3$ input matrix $\shape$ input and the $\numberOfPoints \times \numberOfKeypoints$ output matrix $\hat{\distanceGaussian}$. This naturally fits an encoder-decoder architecture, where we aim to predict features for each point from the input. The point cloud encoder-decoder architectures are usually designed to solve semantic segmentation problems, where the neural network predicts the probability of each point belonging to a specific class. The formulation of an encoder-decoder into a segmentation problem is only enforced by squashing the network prediction into a probability, e.g., usually using a sigmoid function, and by backpropagating the expected class probability through the loss function.

To change the classical segmentation problem into a regression problem, we start by computing the distance on the point cloud manifold\footnote{In the context of a point cloud, the notion of a manifold and geodesic distance can be built by analysing the curvature in local neighbourhoods, e.g., using the Laplace-Beltrami operator.}, $\distanceGeodesic_{i,j} \in \mathbb{R}^{\numberOfPoints \times \numberOfKeypoints}$, between the $i^\text{th}$ point and the $j^\text{th}$ keypoint using the heat kernel method~\cite{Crane:2017:HMD}. The heat kernel method can be implemented on different data structures (e.g., mesh, tetrahedral, point clouds) as long as a Laplacian operator can be computed on this data structure. In our case, we use the \emph{tufted Laplacian}~\cite{Sharp:2020:LNT}, which is robust to noisy data and particularly adapted for point clouds generated by depth cameras.

The computation of the distance on the manifold is computationally expensive and is, therefore, performed as a pre-processing step (i.e., prior to training the network to avoid this computation for all epochs). The geodesic distance is then mapped into a radial basis function (RBF) with a Gaussian kernel to have a localised heat map as follows:
\begin{equation}
    \label{eq:RBF_mapping}
    \distanceGaussian_{i,j} = e^{-\epsilon\left({\distanceGeodesic_{i,j}}\right)^2}
\end{equation}
where $\epsilon$ controls the sharpness of the annotation and should be chosen with respect to the scale of the point cloud (in our case $\epsilon=10$). Mapping the geodesic distance into an RBF serves multiple purposes. Firstly, we believe that it simplifies the learning process by reducing the areas where the distance has to be learned, i.e., only the distance in a local area needs to be learned instead of the full shape. Secondly, it squashes the learned features between $0$ and $1$, allowing us to use a standard mean square error (MSE) loss function as, 
\begin{equation} 
    \label{eq:loss_function}
    l = \frac{1}{\numberOfPoints \numberOfKeypoints}\sum_{i,j}{(\hat{\distanceGaussian}_{i,j}-\distanceGaussian_{i,j})^2},
\end{equation}
without having to modify a network designed for semantic segmentation (i.e., an architecture with an encoder-decoder). Finally, it also allows us to make some specific assumptions in the data augmentation, as discussed in Section~\ref{sec:outputs-data-augmentation}. Given Equation~\eqref{eq:RBF_mapping}, the position of the keypoints is now obtained by finding the argument that maximises the learned distances (i.e., \emph{argmax} of prediction).

In this work, we argue that an encoder-decoder architecture can be used interchangeably as long as it offers the capability to learn on point clouds. Therefore, to demonstrate the flexibility of our approach, in the experiments, we test our method on both Pointnet++~\cite{qi2017pointnet++} and Kp-CONV~\cite{thomas2019kpconv} which are commonly regarded as pillars for learning on point clouds.

\subsection{Network's inputs data augmentation}
\label{sec:cloud-data-augmentation}
The proposed data augmentation considers two main approaches: first, we analyse the data augmentation specific to multi-depth-camera systems, secondly we consider the data augmentation allowing us to augment the training set.


\subsubsection{Multi-depth-camera system augmentation} 
In the literature, many of the public keypoint datasets on point clouds are built by sampling points from meshes. For applications on point clouds obtained from multiple depth cameras, several real-world challenges have to be considered. The camera rig is portable and needs to be mounted, calibrated, and unmounted for each data collection. Therefore, the deep learning model needs to be reliable to a slight change of calibration. This robustness is also important in the case of slight camera motion arising from the vibration of the camera system (in our case, this scenario is particularly relevant due to seldom, though very forceful, kicks from animals onto the structure used for data collection). Data augmentation is formally defined as
\begin{equation}
    \forall \cameraID \in \{1, \dots, \numberOfCameras\}, \shape_\cameraID = \shape_\cameraID \boldsymbol{R} + \boldsymbol{t}
\end{equation}
with $\boldsymbol{R}$ a rotation matrix in $SO(3)$ and $\boldsymbol{t}$ a translation vector that add calibration noise.

Additionally to enforce robustness to calibration errors, the camera frames dropout has to be considered. This can occur due to network traffic between computers in the system or latency in data acquisition due to the depth camera's auto-exposure. This problem has been studied in the context of multi-view CNN by embedding a \emph{dropview} layer within the architecture at training time~\cite{huang2019deepccfv}. This can be adapted for encoder-decoder by dropping some of the point clouds from P such as :
\begin{equation}
    \label{eq:camera_dropout}
    \shape = \bigcup\limits_{\cameraID=1}^{\numberOfCameras}
    \begin{cases}
        \shape_\cameraID  & \text{if } x\leq \cameraDropProbability \mid x \sim U([0,1])\\
        \emptyset & \text{otherwise}
    \end{cases}
\end{equation}
where $x$ is the probability of dropping the camera point cloud. Keeping the point cloud dropout within the data loader as part of the data augmentation allows our method to stay agnostic to the model architecture.

\subsubsection{Geometric transformation}
\label{sec:geometric-transformation}
Similarly to~\cite{qi2017pointnet++, qi2019deep}, we propose to use changes of scale which can be formulated as a geometric transformation. We also consider random flipping similarly to~\cite{qi2018frustum, qi2019deep}. Following~\cite[Sec.~4.7.6]{schneider2002geometric}, the scale and random flipping transformation matrices are defined as:
\begin{equation}
    \Tscale =   \begin{bmatrix}
                s_x & 0   & 0   & 0\\
                0   & s_y & 0   & 0\\
                0   & 0   & s_z & 0\\
                0   & 0   & 0   & 1
                \end{bmatrix},
    \Tflip =   \begin{bmatrix}
                f & 0 & 0 & 0\\
                0 & 1 & 0 & 0\\
                0 & 0 & 1 & 0\\
                0 & 0 & 0 & 1
                \end{bmatrix}
\end{equation}
where $s_x$, $s_y$, and $s_z$ are drawn from a normal distribution $\mathcal{N}(1, 0.1)$. $f$ corresponds to a random reflection of the shape with respect to the YZ symmetry plane, such as:
\begin{equation}
    f = \begin{cases}
  -1, & \text{if } x > 0.5 \mid x \sim U([0,1])\\
  1, & \text{otherwise}.
\end{cases}
\end{equation}

Additional augmentation methods can be obtained using geometric transformations such as translation and rotations. The networks for point clouds usually learn on structures defined from local neighbourhoods and are therefore invariant to global translation. Rotation transformations have been used in the literature~\cite{qi2017pointnet, qi2017pointnet++, qi2019deep}. However, in our case the dedicated perception system scans animals in a similar orientation as shown in Figure~\ref{fig:data_collection}, therefore random rotations are out of the scope of our work. 

\subsubsection{Non-rigid deformation} 
Having control of the animal stance, i.e., how they stand during the data collection, is extremely challenging. As an alternative, one would rather learn to perform the keypoints annotation for all possible animal poses. This would ideally be obtained through the application of non-rigid deformation of the animal shape by simulating the motion of the animal.

In the field of non-rigid deformation, this is typically performed by rigging the shape of the animal (building a virtual skeleton) and animating its motion~\cite{kavan2008geometric, jacobson2011bounded}. In our case, such deformation is unobtainable as the virtual skeleton is unavailable. Further, this is to some extent what this paper is aiming to solve (with the difference that we are aiming at finding points on the point cloud rather than nodes from the skeleton). Alternatively, 3D shapes can be deformed without a virtual skeleton as a prior by maximising the local rigidity of the shape while trying to minimise the error on ``handles'' in motion (handles are points selected from the shape). While near real-time implementations of such methods exist for meshes~\cite{sorkine2007rigid, jacobson2012fast}, the equivalent for point clouds are significantly slower~\cite{sumner2007embedded}. Regardless of the computation time, the deformation requires the non-rigid motion to be parameterised, which is also challenging.


As an alternative to these traditional methods, we propose to use shear matrices~\cite[Sec.~4.7.6]{schneider2002geometric} to simulate the motion of the animals stepping sideways and leaning forward / backward. These shear matrices are defined as:
\begin{equation}
    \Tshearside =    \begin{bmatrix}
                     1 & 0 & 0 & 0\\
                     h_s & 1 & 0 & 0\\
                     0 & 0 & 1 & 0\\
                     0 & 0 & 0 & 1
                     \end{bmatrix}, 
    \Tshearforward = \begin{bmatrix}
                     1 & 0 & 0 & 0\\
                     0 & 1 & h_f & 0\\
                     0 & 0 & 1 & 0\\
                     0 & 0 & 0 & 1
                     \end{bmatrix}
\end{equation}
where $h_s = \tan(\theta_s)$ and $\theta_s \sim \mathcal{N}(0, \frac{\pi}{20})$ the shearing angle with respect to the $x$ axis. Similarly, $h_f = \tan(\theta_f)$ and $\theta_f \sim \mathcal{N}(0, \frac{\pi}{20})$ the shearing angle with respect to the $z$ axis. Formulating the motion as additional transformation matrices allows us to obtain a non-rigid deformation matrix that can be easily parameterised and applied simultaneously as to the other data-augmentation discussed in Section~\ref{sec:geometric-transformation}.

\begin{figure}
\begin{tabular}{@{}C{0.06\linewidth}@{ }C{0.3\linewidth}@{ }C{0.3\linewidth}@{ }C{0.3\linewidth}@{ }@{ }@{}}
    & \vcentered{I} & \vcentered{II} & \vcentered{III}\\
    (a) &
    \vcentered{\includegraphics[width=\linewidth]{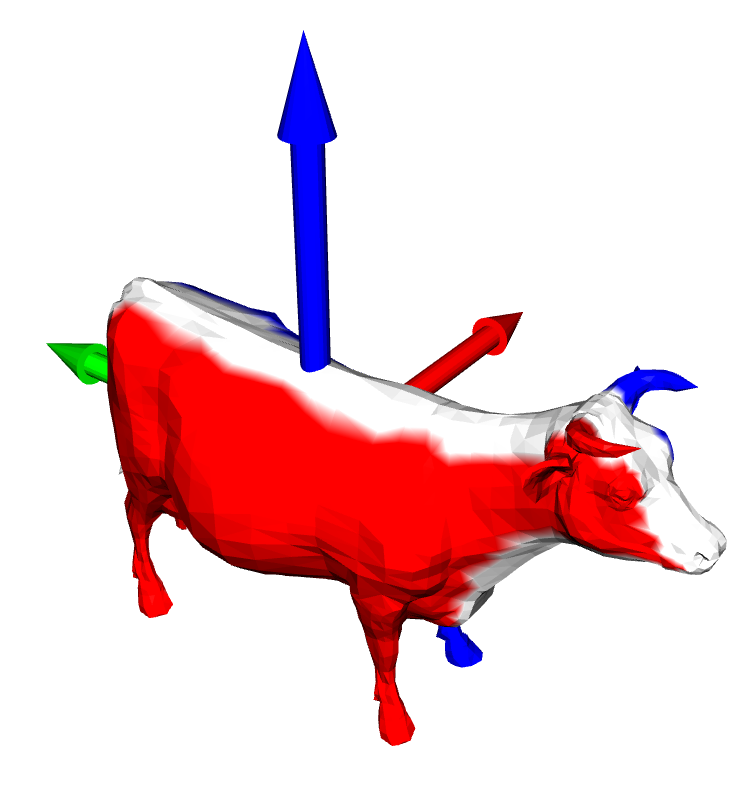}}&
    \vcentered{\includegraphics[width=\linewidth]{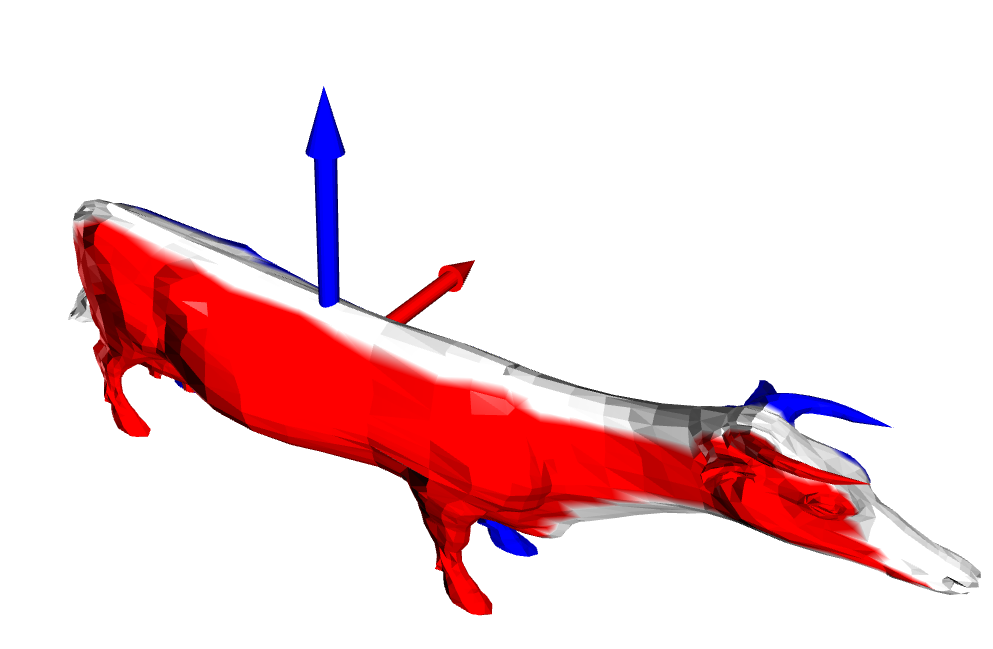}}&
    \vcentered{\includegraphics[width=\linewidth]{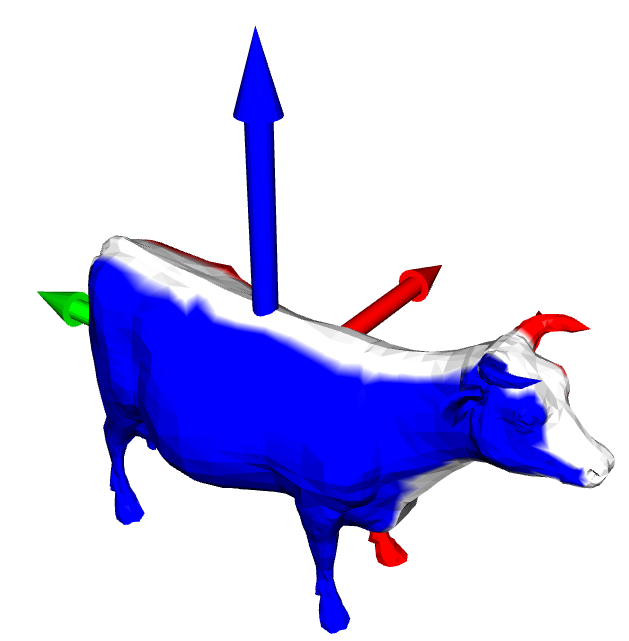}}\\
    (b)& &
    \vcentered{\includegraphics[width=\linewidth]{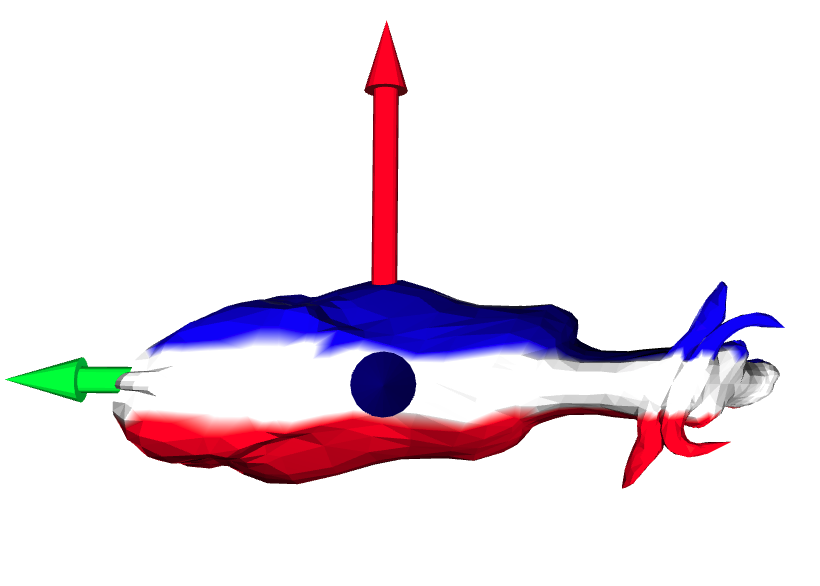}}&
    \vcentered{\includegraphics[width=\linewidth]{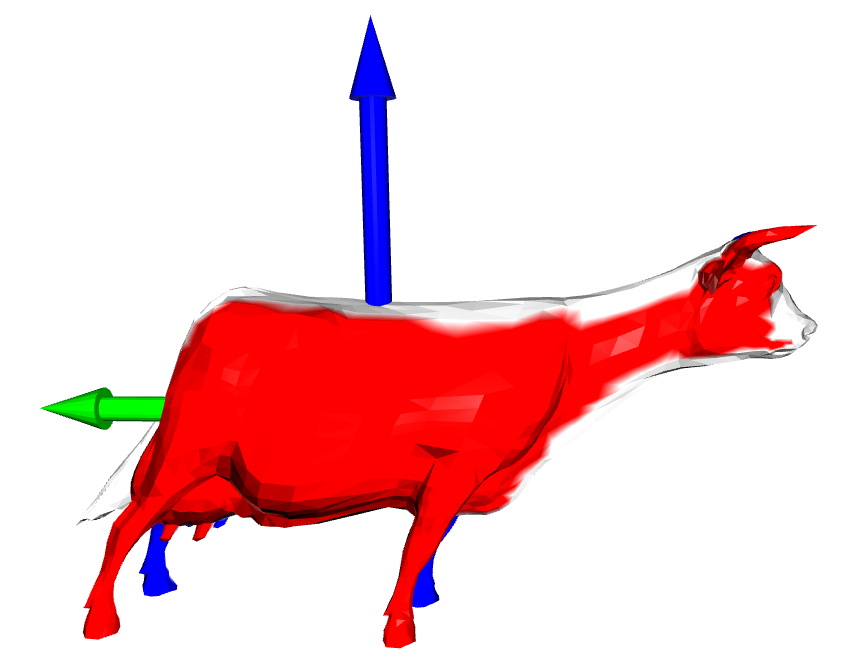}}\\
\end{tabular}
\caption{Data augmentation applied to a mesh where the parameters are amplified for visualisation. The coordinate system is defined by the $x$, $y$, and $z$ axes respectively displayed as red, green, and blue arrows. The original shape in I.(a), the forward shearing in II.(a), the side-way shearing in III.(a), the non-isotropic scaling in II.(b), and the random flipping in III.(b).}
\label{fig:data-augmentation}
\end{figure}

\subsubsection{Application of the data augmentation}
Given all the different geometric transformation matrices, we can apply this transformation to the points of the shape and the annotated keypoints as follow:
\begin{align}
\label{eq:transformation}
    \shapeUpdated &= \Tscale \Tflip \Tshearside \Tshearforward \shape\\
    \keypointsUpdated &= \Tscale \Tflip \Tshearside \Tshearforward \keypoints
\end{align}
A visualisation of the cloud data augmentation is proposed in Figure~\ref{fig:data-augmentation}.

\subsection{Network's outputs data augmentation}
\label{sec:outputs-data-augmentation}
In the case of semantic segmentation, the data augmentation rarely impacts the networks outputs and is often overlooked in the literature (e.g., for semantic segmentation, points labels are constant). However, in the case of the regression of the distance on the manifold, the input non-rigid deformation (i.e., $\Tscale$, $\Tshearforward$, and $\Tshearside$) should change the predicted distances and the data augmentation has to be considered during the loss function computation. Ideally, the distance on the manifold with respect to the annotated point would be recomputed. However, given the computational complexity of calculating the distance on the manifold, we propose to update the distance on the manifold with respect to the difference in the Euclidean distance between $\shape$ and $\shapeUpdated$ from Equation~\eqref{eq:transformation}. More formally, the distance on the manifold is first approximated as
\begin{equation} 
    \label{eq:updated_geodesic}
    \distanceGeodesicUpdated_{i,j} = \distanceGeodesic_{i,j} + \distanceEuclideanUpdated_{i,j} - \distanceEuclidean_{i,j},
\end{equation}
and the RBF distances are updated as follow,
\begin{equation}
    \label{eq:updated_RBF}
    \distanceGaussianUpdated_{i,j} = e^{-\epsilon\left({\distanceGeodesicUpdated_{i,j}}\right)^2}
\end{equation}
where $\distanceEuclidean =  d(\shape, \keypoints)$ and $\distanceEuclideanUpdated = d(\shapeUpdated, \keypointsUpdated)$ are the Euclidean distances before and after equation \eqref{eq:transformation}. $\distanceGaussianUpdated$ is the updated distance on the manifold mapped into an RBF.

The case of the flip transform, $\Tflip$, has to be handled differently. The transformation applies a symmetry on the YZ plane to the shape and therefore, it changes the semantic meaning of the annotation (e.g., the left legs become the right legs within the transformation). To update the outputs, a simple reshuffling of the distance vector to the annotation is necessary. Using pseudo-code with 0 indexing, the update of the outputs is reshuffled as
\begin{equation} 
    \distanceGaussianUpdated = \distanceGaussianUpdated[:, [5, 4, 2, 3, 1, 0]].
\end{equation}

Finally, following the camera frame dropout implemented in Equation~\eqref{eq:camera_dropout}, $\distanceGaussianUpdated$ is updated to drop the element with indexes corresponding to the dropped cameras.

\begin{figure}
    \centering
    \includegraphics[width=0.95\linewidth]{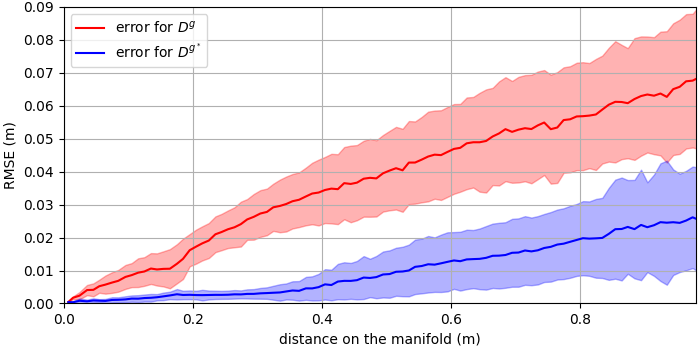}
    \caption{Mean error and standard deviation for $\distanceGeodesic$ and $\distanceGeodesicUpdated$ versus the recomputed geodesic distance following \eqref{eq:transformation}. As shown in the plot, while the approximation considerably reduce the error, our approximation is only valid in the vicinity of the keypoints. Therefore, the use of the RBF mapping in~\eqref{eq:updated_RBF} makes this approximation acceptable.}
    \label{fig:}
\end{figure}


\section{Experimental Results}
\label{sec:result}

\begin{figure}
	\centering
	\includegraphics[height=2.5cm]{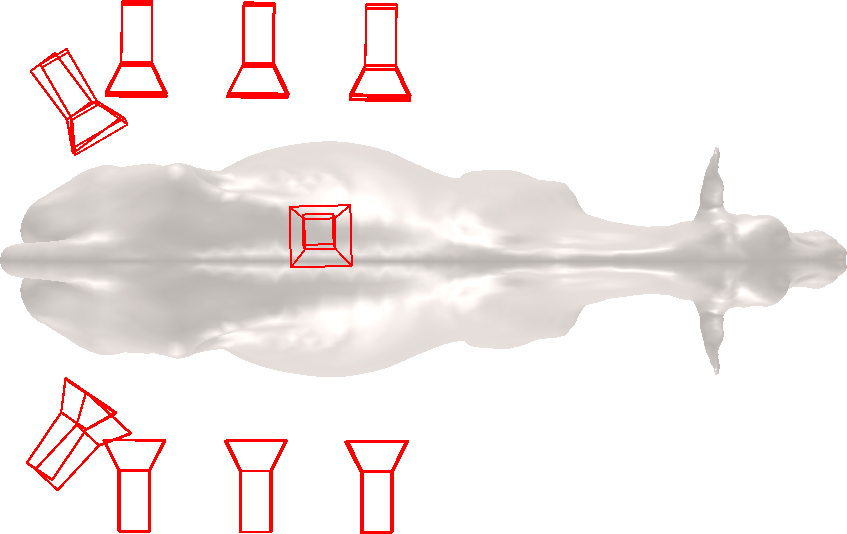}
	\includegraphics[height=2.5cm]{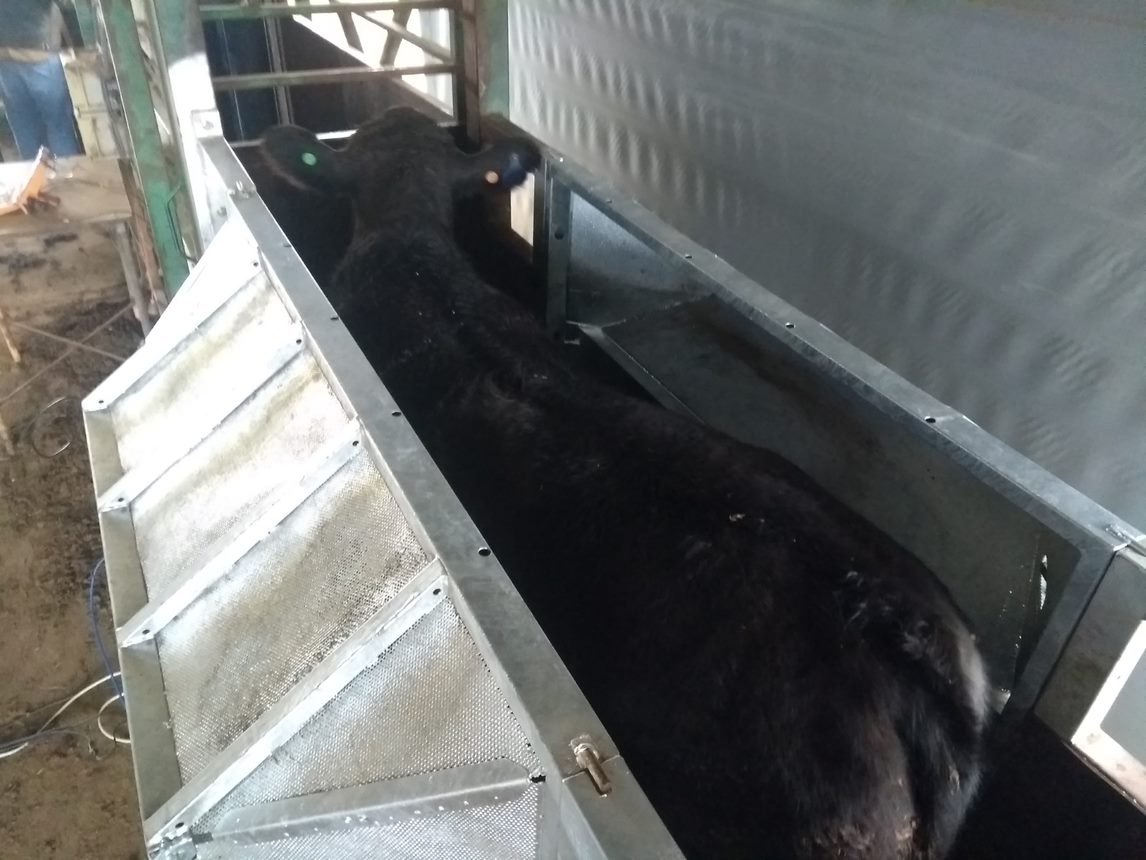}
	\caption{Data collection system used to scan cattle: the perception system is made of multiple RGB-D realsense cameras that are positioned to  allow capturing the full shape of the animals in a single snapshot (each side of the camera has two rows of four cameras and one camera is located above). Overlap between the cameras allows for extrinsic calibration and ensure that the data collection is more reliable.}
	\label{fig:data_collection}
\end{figure}

To demonstrate the validity of the approach, we use real data taken in a cattle research facility with an in-house built system.
The data collection has been performed at the Tullimba feedlot research facility, where a race (i.e., corridor cattle walk through) has been modified to integrate a robotics system capable of scanning cattle, as shown in Figure~\ref{fig:data_collection}. The robotics perception system is designed with 17 RGB-D Realsense D435 cameras extrinsically calibrated with a checkerboard prior to the data collection. Cameras are synchronised using a hardware trigger allowing to capture an animal's 3D shape in a single snapshot. The data collection was undertaken on two different occasions, with 274 animals scanned on the first date and a different cohort of 136 animals scanned on the second date. On each occasion, the system is assembled and calibrated. Therefore, we consider the data collections as two independent dataset with the larger one used as training set and the smaller one used as testing set. This allows us to test if the model overfits a specific dataset or if they can be used across different datasets regardless of the calibration.

For each animals, the point clouds are stitched together using the extrinsic calibration. The background is removed to isolate the animal, and outliers are filtered using radius outlier removal (ROR) and statistical outlier removal (SOR) filters~\cite{rusu2010semantic} resulting in $\shape$. All the point clouds from both datasets have then been annotated manually with 6 keypoints (the top of each legs, the hip, and the neck), and the geodesic distance~\cite{Crane:2017:HMD} between the annotated keypoints and the rest of the point cloud has been pre-computed. As a result, every instance of our dataset can been considered as a tuple of three matrices $\left\{\shape, \keypoints, \distanceGeodesic\right\}$.

For the training, we investigated both the Pointnet++ and Kp-CONV architectures while using standard tools and default hyper-parameters during the training stage. While this is not the focus of the paper, we provide the hyper-parameters for the sake of algorithmic reproducibility. The optimisation of the network parameters has been done using stochastic gradient descent (SGD) with a learning rate of $0.01$ using a momentum of $0.9$~\footnote{The learning rate for Kp-CONV could have been increased as it learned significantly slower but smoother than Pointnet++.}.

The error on the Euclidean distance between the annotated points and the points predicted by both networks is reported in Table~\ref{tab:keypoint_distance} and a study on the sampling size of the training set is given in Table~\ref{tab:sampling_study}. The data augmentation allows to maintain good performances while being able to learn on a relatively small dataset. It should be noted that the remaining error does not account for the error of manual annotations.

\begin{figure}
    \centering
    \includegraphics[width=0.49\linewidth]{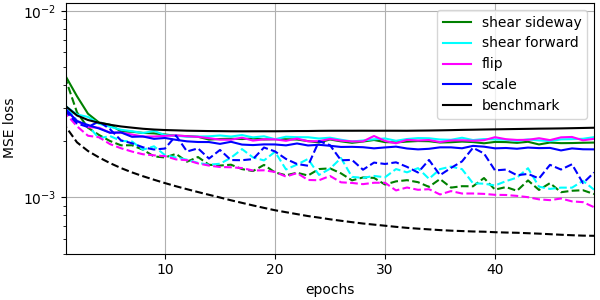}
    \includegraphics[width=0.49\linewidth]{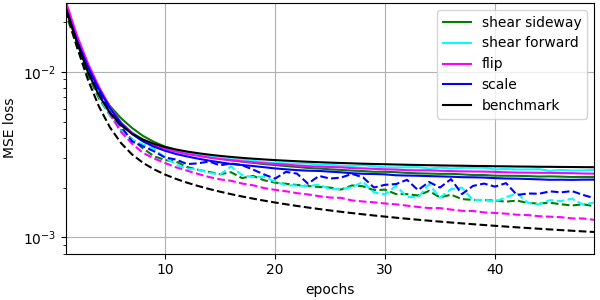}
    \includegraphics[width=0.49\linewidth]{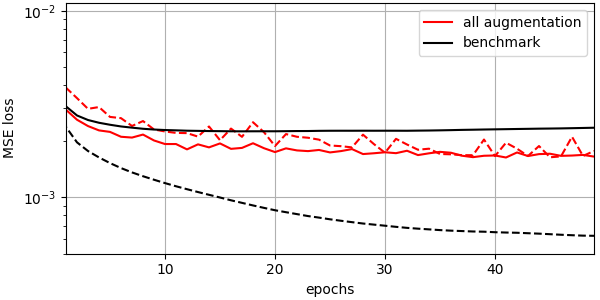}
    \includegraphics[width=0.49\linewidth]{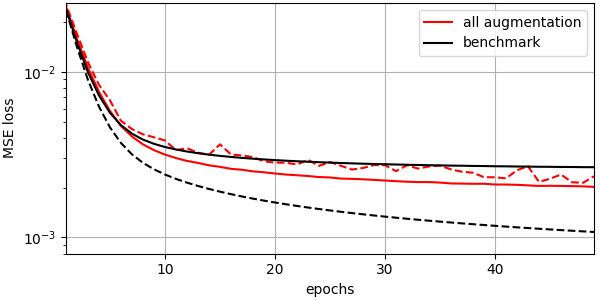}
    \caption{Training loss (dashed) and testing loss (plain) in a log scale versus the number of epochs for the Pointnet++~\cite{qi2017pointnet++} (left) and Kp-CONV~\cite{thomas2019kpconv} (right) networks. The top graphs are an ablation study for each data augmentation, showing they reduce the gap between training and testing loss functions while improving the loss for the testing set when compared to the benchmark with no data augmentation (in black). Once combined, the data augmentation avoids overfitting the training set which can be seen as the loss function on the testing set is bellow the training set. This is particularly relevant for the Pointnet++ architecture, where the networks stops learning after 20 epochs.}
    \label{fig:data_augmentation_ablation_study}
\end{figure}

\begin{table}
    \centering
    \caption{RMSE of the keypoint prediction (in cm) for Pointnet++ and Kp-Conv with/without data augmentation. The error decreases significantly with data augmentation. To put these results in perspective, the bounding box diagonal of a cow scan would be approximately 2m.}
    \begin{tabular}{ c c c c }
     \toprule
               &  w.o. augmentation &  w. augmentation &  w. augmentation \\ 
               & (50 epochs) & (50 epochs) & (150 epochs) \\
     \midrule
     Pointnet++  &  7.02    &  5.76              &  5.60                          \\ 
     Kp-CONV   &  7.15    &  6.41              &  5.98                          \\ 
     \bottomrule
    \end{tabular}
    \label{tab:keypoint_distance}
\end{table}

\begin{table}
    \centering
    \caption{Study of the data augmentation effectiveness with respect to the training set size. The error reported are the MSE (at the power $10^{-3}$) of the updated distance on the manifold on the testing set (see Equation~\eqref{eq:loss_function}). The results show that the data augmentation allows to avoid a drop of performance with a reduced training set.}
    \begin{tabular}{ c c c c c }
     \toprule
               &  \multicolumn{2}{c}{Pointnet++} & \multicolumn{2}{c}{Kp-CONV} \\ 
     \toprule
     augmentation &  with &  without &  with & without  \\
     \midrule
     200 samples  &  0.90    &  1.01  &  1.12 &  1.32 \\ 
     100 samples  &  0.88    &  1.31  &  1.17 &  1.50 \\ 
     50 samples   &  0.94    &  1.41  &  1.13 &  1.76 \\ 
     \bottomrule
    \end{tabular}
    \label{tab:sampling_study}
\end{table}

\begin{figure}
	\centering
		\includegraphics[height=3.6cm]{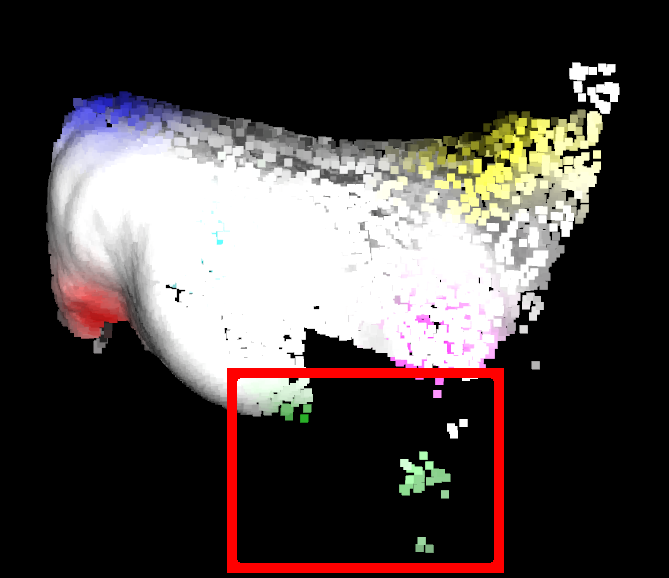}
		\includegraphics[height=3.6cm]{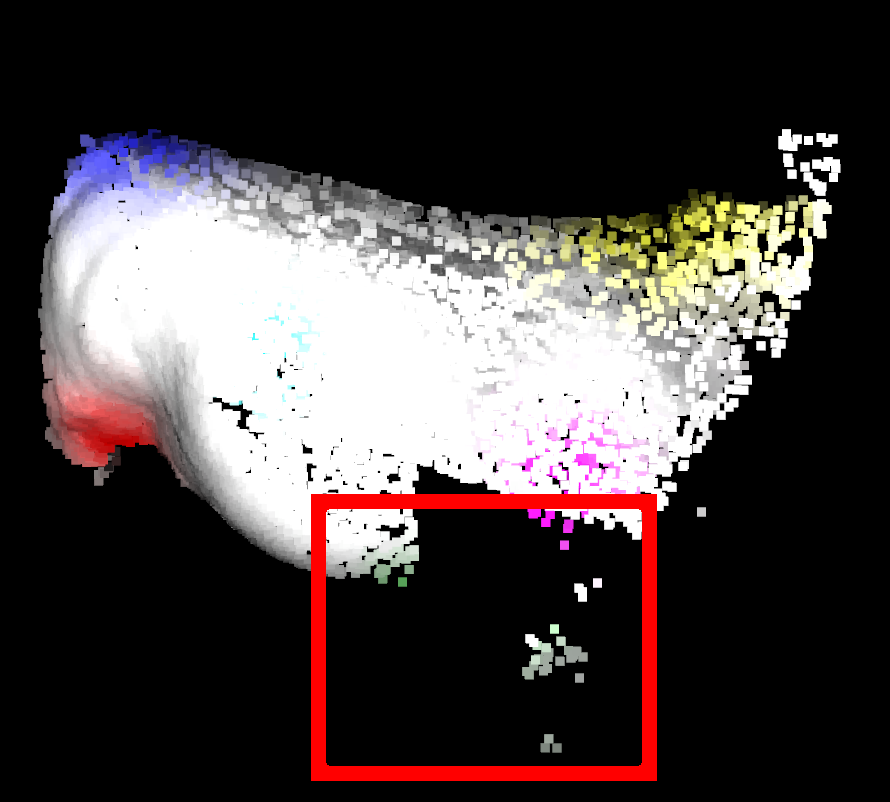}
	\caption{Sample of point cloud with dropped frame. The camera frame dropout in the data augmentation allows avoiding being overconfident in the prediction of the green keypoints. The maximum value for the predicted distance on the manifold drops from 0.79 to 0.45. As a result, we can then use the predicted distance on the manifold as a measure of the keypoints uncertainty.}
	\label{fig:camera_frame_dropout}
\end{figure}

\begin{figure}
\centering
\begin{tabular}{@{}C{0.04\linewidth}@{ }C{0.31\linewidth}@{ }C{0.31\linewidth}@{ }C{0.31\linewidth}@{ }@{ }@{}}
& \vcentered{manual} & \vcentered{pointnet++} & \vcentered{Kp-CONV} \\
\vcentered{(a)}&
\vcentered{\includegraphics[width=\linewidth]{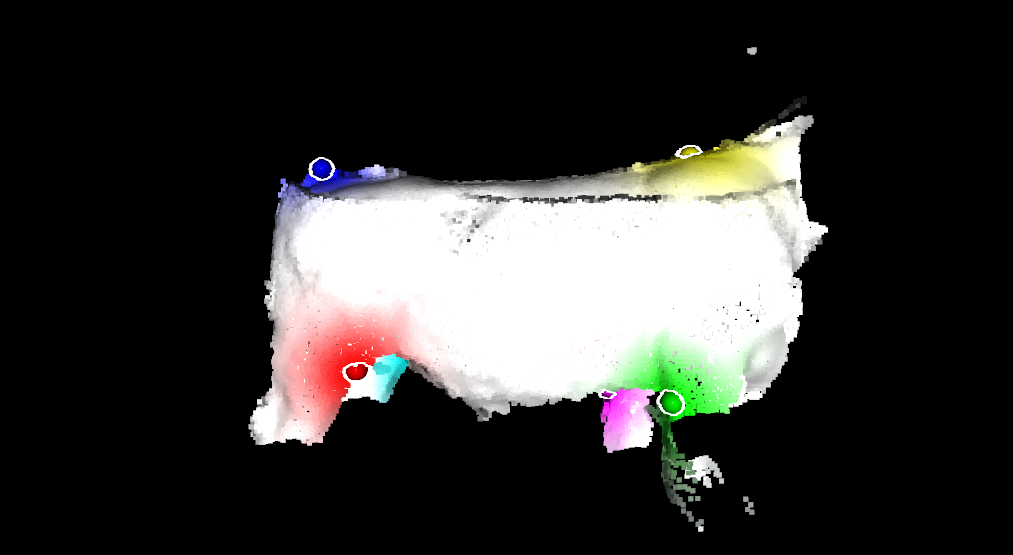}}&
\vcentered{\includegraphics[width=\linewidth]{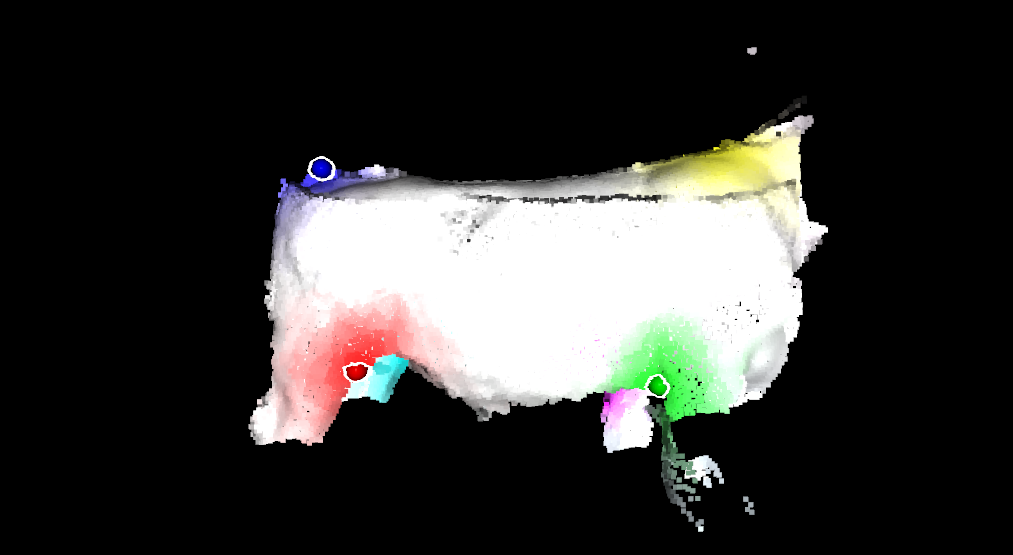}}&
\vcentered{\includegraphics[width=\linewidth]{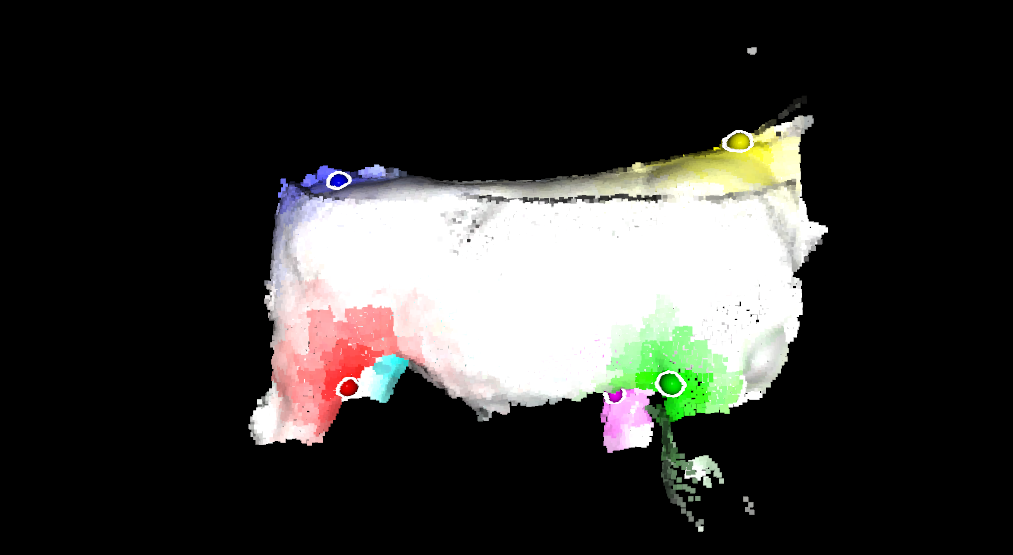}}\\

\vcentered{(b)}&
\vcentered{\includegraphics[width=\linewidth]{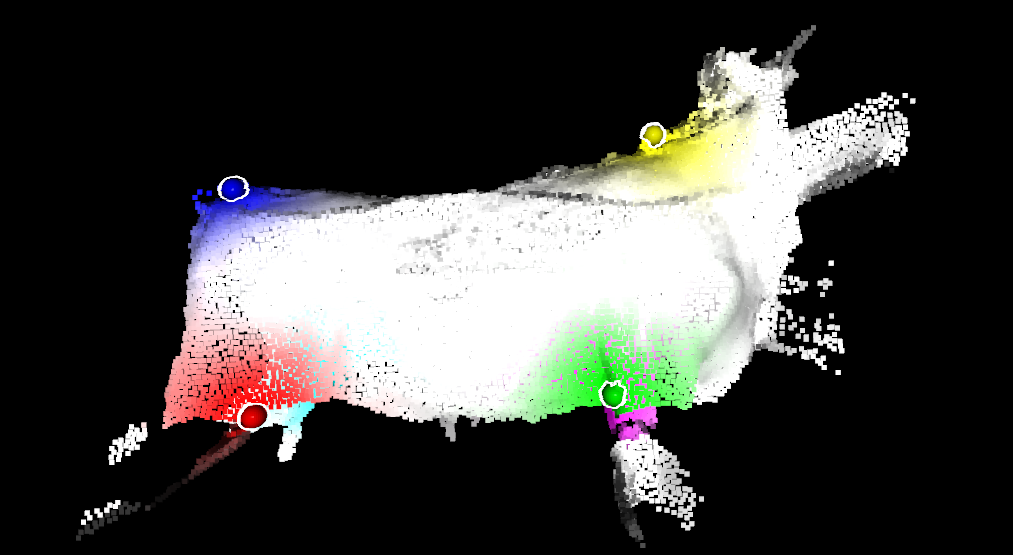}}&
\vcentered{\includegraphics[width=\linewidth]{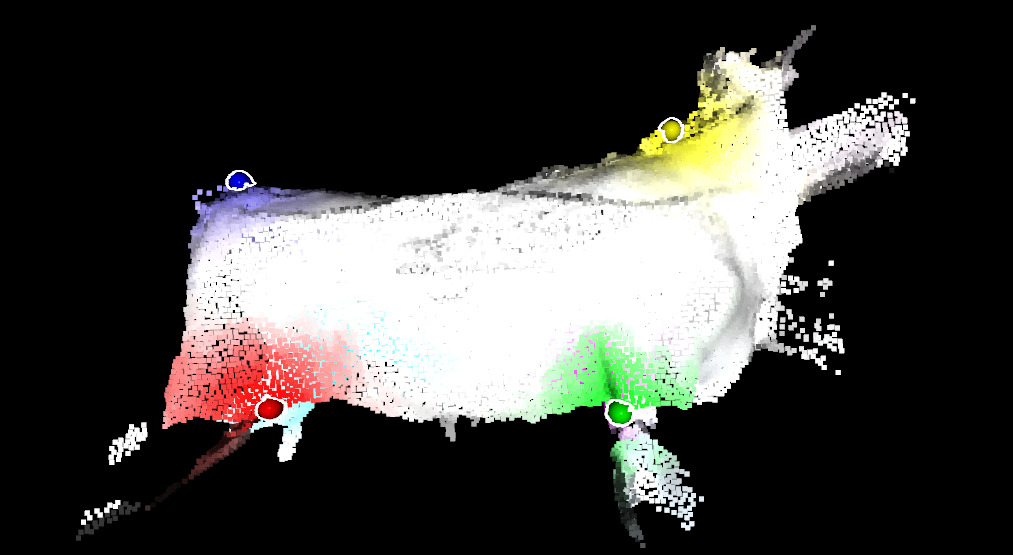}}&
\vcentered{\includegraphics[width=\linewidth]{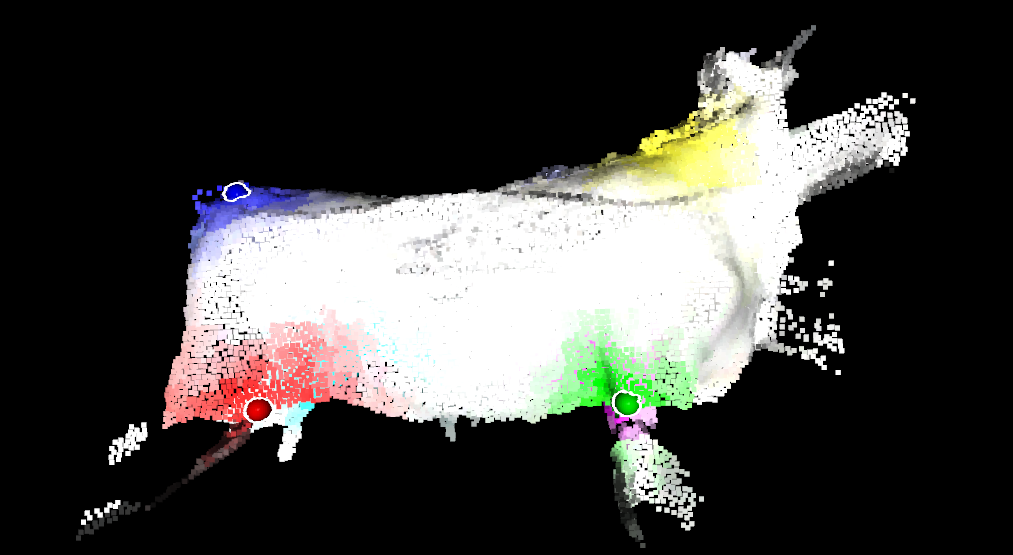}}\\

\vcentered{(c)}&
\vcentered{\includegraphics[width=\linewidth]{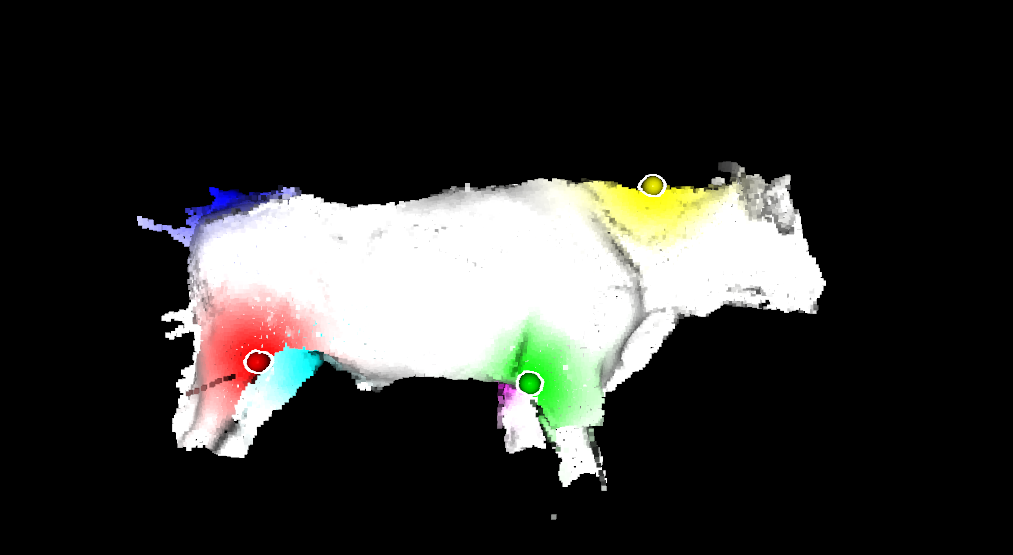}}&
\vcentered{\includegraphics[width=\linewidth]{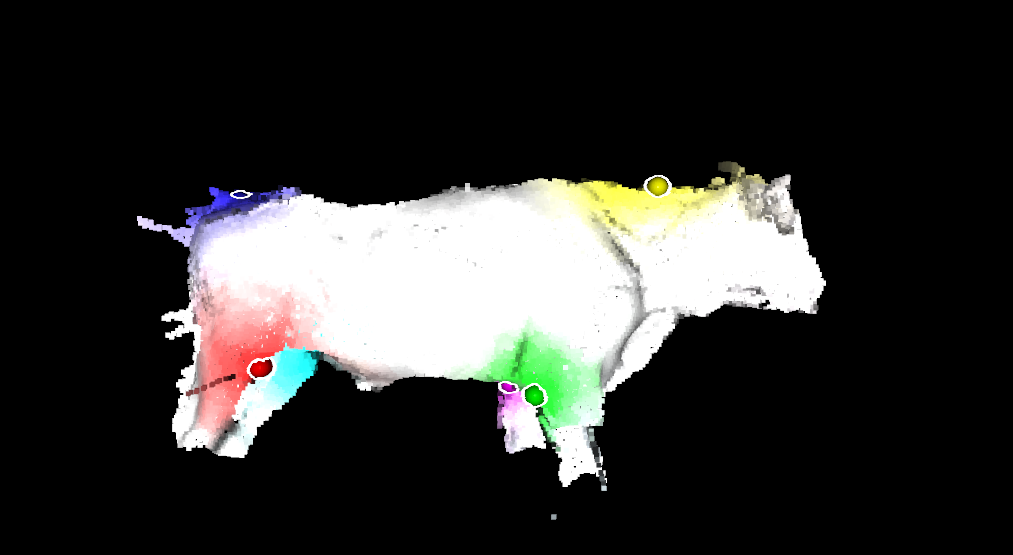}}&
\vcentered{\includegraphics[width=\linewidth]{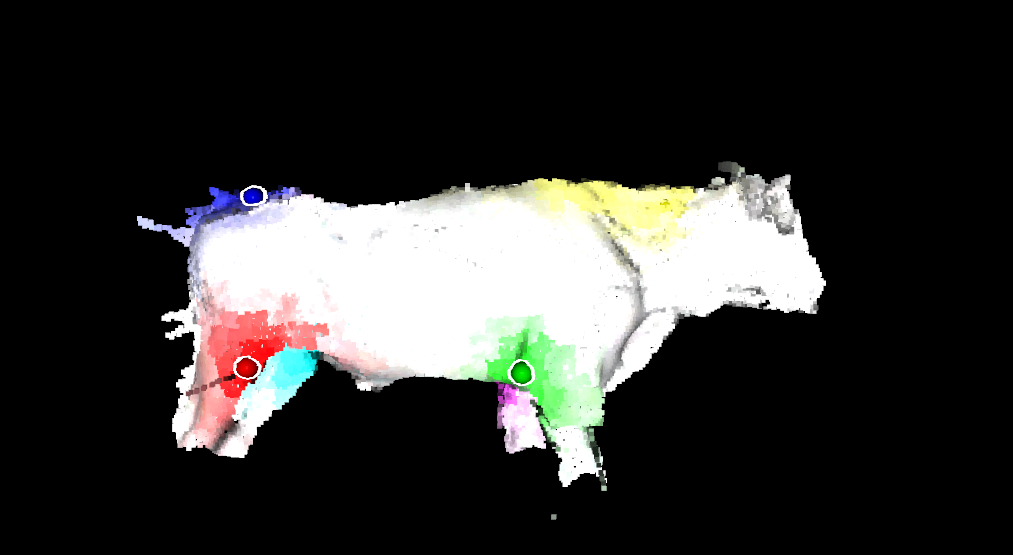}}\\

\vcentered{(d)}&
\vcentered{\includegraphics[width=\linewidth]{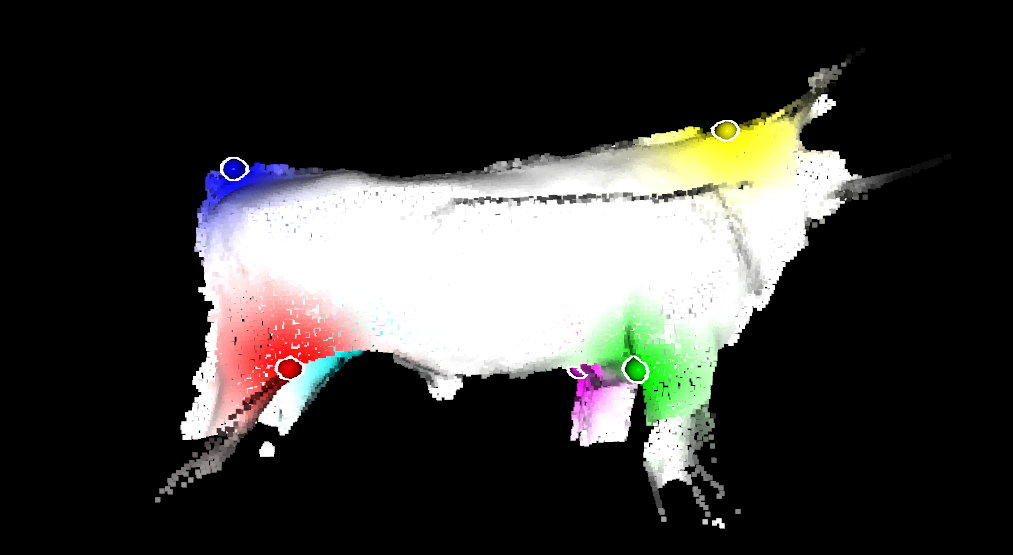}}&
\vcentered{\includegraphics[width=\linewidth]{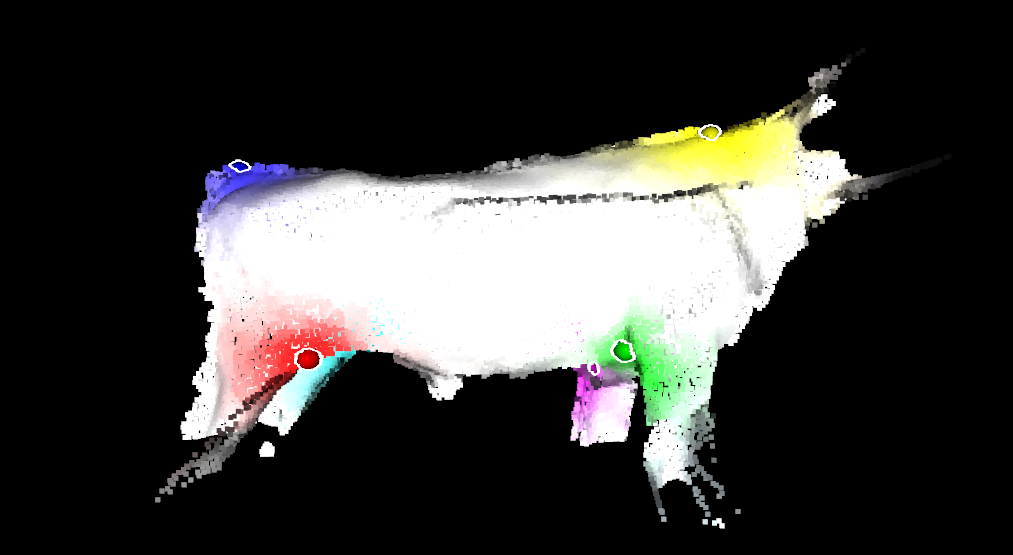}}&
\vcentered{\includegraphics[width=\linewidth]{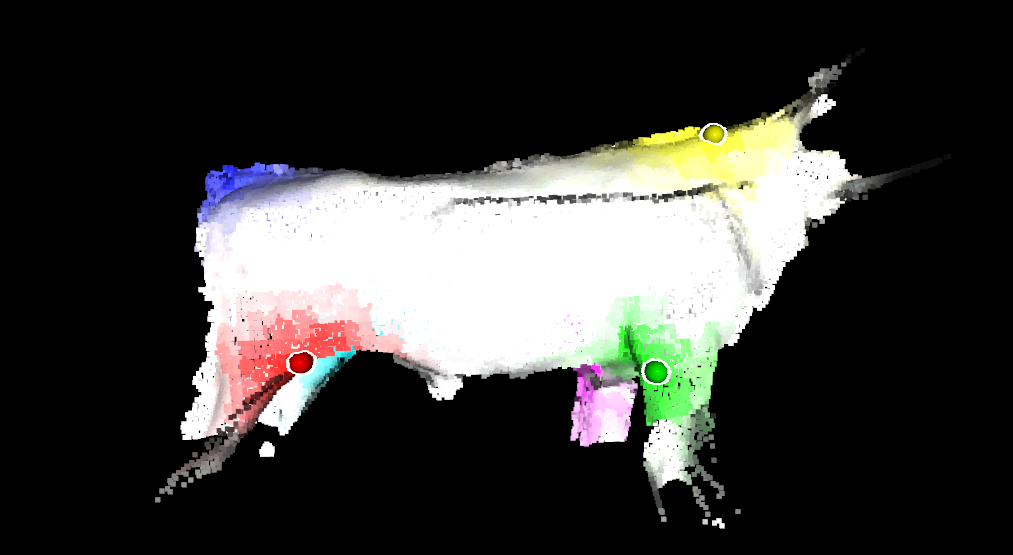}}\\
\end{tabular}
\caption{Keypoints manually annotated, and predicted by Pointnet++ and Kp-CONV are highlighted in white. The cow's pointclouds are coloured with the distance on the manifold to each keypoints. In (a) and (b), the two first samples from the testing set and in (c) and (d) the worst performing predictions. The keypoints with the largest error are the neck which is harder to annotate manually consistently.}
\vspace{-0.6cm}
\end{figure}

In Figure~\ref{fig:data_augmentation_ablation_study}, we perform an ablation study of the data augmentation and show that every non-rigid deformation data augmentation reduce the gap between the training and testing set. When we combine all the augmentation methods from Section~\ref{sec:geometric-transformation}, we can see that both networks do not overfit the training set in the first 50 epochs (which they would most likely do after a certain quantity of epochs). This is a significant improvement when compared to the case without data augmentation (in black), particularly for Pointnet++, which starts overfitting the training set after the first epoch.

A study case of the camera dropout is shown in Figure~\ref{fig:camera_frame_dropout}. The distance prediction from Pointnet++ trained without camera frame dropout (shown in the left) is overconfident in the distance measure. After including the camera frame dropout (shown on the right), the maximum distance measure drops from 0.79 to 0.45, showing that the keypoint is far from the argmax of the prediction. Therefore, the predicted distance of the manifold can be used as a quantification of the keypoints uncertainty.

\section{Conclusion}
\label{sec:conclusion}
In this paper, we propose a simple yet effective method for annotating keypoints using a supervised deep learning approach. We reformulate the problem of keypoints extraction into a regression problem by inferring the distance on the manifold to keypoints. We study the relevant data augmentation methods for multi-depth-camera systems and efficient non-rigid deformation methods. Our method shows relatively good results when compared to the size of an animal's complete scan while being trained on a relatively small dataset. The experiments show that the model can be trained on a relatively small dataset while being robust to camera drops and calibration noise.

Our approach uses a single set of manual annotations for each point clouds, the error on human annotation could be mitigated by using a consensus of expert annotations similarly to~\cite{you2020keypointnet}.


\section*{Acknowledgment}
This paper is supported by funding from the Australian Government Department of Agriculture and Water Resources as part of its Rural R\&D for Profit program, MLA grant number {V.RDP.2005}.


\bibliography{bibliography}  
\bibliographystyle{IEEEtran}

\end{document}